\documentclass[10pt,twocolumn,letterpaper]{article}

\usepackage{cvpr}              % To produce the CAMERA-READY version
\makeatletter
\@namedef{ver@everyshi.sty}{}
\makeatother
\usepackage{everyshi}

\usepackage{graphicx}
\usepackage{amsmath}
\usepackage{amssymb}
\usepackage{booktabs}

\usepackage[pagebackref,breaklinks,colorlinks]{hyperref}

\usepackage{amsmath,amssymb}

\usepackage[dvipsnames]{xcolor}

\usepackage{amsmath}
\usepackage{amssymb}
\usepackage{mathtools}

\usepackage{hyperref}
\usepackage{cleveref}

\usepackage{annotate-equations}

\usepackage[skins,listings]{tcolorbox}
\newtcblisting{LTXexample}[1][]{%
  colframe=black!20,
  colback=black!10,
  coltitle=red!50!yellow!3!white,
  bicolor,colbacklower=white,
  fonttitle=\sffamily\bfseries,
  text above listing,
  #1}

\usepackage{multirow,tabularx}

\usepackage[ruled,vlined]{algorithm2e}

\definecolor{commentcolor}{RGB}{110,154,155}   % define comment color
\newcommand{\PyComment}[1]{\ttfamily\textcolor{commentcolor}{\# #1}}  % add a "#" before the input text "#1"
\newcommand{\PyCode}[1]{\ttfamily\textcolor{black}{#1}} % \ttfamily is the code font

\newcommand{\cL}{\mathcal{L}}

\newcommand{\rI}{\mathrm{I}}
\newcommand{\rT}{\mathrm{T}}
\newcommand{\on}{\overline{n}}

\newcommand{\bigO}{\mathcal{O}}

\def\cvprPaperID{9072} % *** Enter the CVPR Paper ID here
\def\confName{CVPR}
\def\confYear{2023}

\begin{document}

%%%%%%%%% TITLE - PLEASE UPDATE
\title{DisCo-CLIP: A Distributed Contrastive Loss \\ for Memory Efficient CLIP Training}
%\title{ME-CLIP: Memory Efficient CLIP Training}

\author{Yihao Chen, Xianbiao Qi, 
Jianan Wang, Lei Zhang\thanks{Corresponding author.} \\
International Digital Economy Academy (IDEA), Shenzhen, Guangdong, China.\\
\texttt{\{chenyihao, qixianbiao, wangjianan, leizhang\}@idea.edu.cn} 
}

%\author{Yihao Chen}
%, Xianbiao Qi, Jianan Wang, Lei Zhang \\
%Institution1\\
%Institution1 address\\
%{\tt\small firstauthor@i1.org}
% For a paper whose authors are all at the same institution,
% omit the following lines up until the closing ``}''.
% Additional authors and addresses can be added with ``\and'',
% just like the second author.
% To save space, use either the email address or home page, not both

%}
\maketitle

%\and
%Second Author\\
%Institution2\\
%First line of institution2 address\\
%{\tt\small secondauthor@i2.org}

%%%%%%%%% ABSTRACT
\begin{abstract}
   We propose DisCo-CLIP, a distributed memory-efficient CLIP training approach, to reduce the memory consumption of contrastive loss when training contrastive learning models. Our approach decomposes the contrastive loss and its gradient computation into two parts, one to calculate the intra-GPU gradients and the other to compute the inter-GPU gradients. According to our decomposition, only the intra-GPU gradients are computed on the current GPU, while the inter-GPU gradients are collected via \texttt{all\_reduce} from other GPUs instead of being repeatedly computed on every GPU. In this way, we can reduce the GPU memory consumption of contrastive loss computation from $\bigO(B^2)$ to $\bigO(\frac{B^2}{N})$, where $B$ and $N$ are the batch size and the number of GPUs used for training. Such a distributed solution is mathematically equivalent to the original non-distributed contrastive loss computation, without sacrificing any computation accuracy. It is particularly efficient for large-batch CLIP training. For instance, DisCo-CLIP can enable contrastive training of a ViT-B/32 model with a batch size of 32K or 196K using 8 or 64 A100 40GB GPUs, compared with the original CLIP solution which requires 128 A100 40GB GPUs to train a ViT-B/32 model with a batch size of 32K. The code will be released at \url{https://github.com/IDEA-Research/DisCo-CLIP}
\end{abstract}

\section{Introduction}
Vision-language representation learning from massive image-text pairs has recently attracted tremendous attention for its great potential in many applications such as zero-shot classification and text-image retrieval. Representative works include CLIP~\cite{clip_radford2021learning}, ALIGN~\cite{align_jia2021scaling}, Florence~\cite{florence_yuan2021florence}, CoCa~\cite{coca_yu2022coca}, and BASIC~\cite{basic_pham2021combined}, which all leverage hundreds of millions or even billions of image-text pairs collected from the Web to learn a semantic-rich and language-aligned visual representation~\cite{glip_li2022grounded}. As the web-collected data inevitably contain noises, CLIP~\cite{clip_radford2021learning} for the first time applies contrastive learning on 400M image-text pairs, which implies a weak but more proper assumption about the data: the relevance between paired image and text is greater than that between unpaired image and text. For its demonstrated performance in CLIP, contrastive learning has been widely adopted in subsequent works. Accordingly, several image-text data sets with increasingly larger scales have also been developed and made publicly available, such as Conceptual 12M~\cite{cc12_changpinyo2021conceptual}, YFCC 100M~\cite{yfcc_thomee2016yfcc100m}, WIT 37.6M~\cite{wit_srinivasan2021wit}, LAION-400M~\cite{laion400_schuhmann2021laion}, and LAION-5B~\cite{laion5b_schuhmann2022laion}.

The goal of contrastive learning in CLIP is to learn an alignment between image and text via two encoders. That is, it encourages paired image and text (called a positive pair) to be similar and meanwhile enforces unpaired image and text (called a negative pair) to be dissimilar. For any positive image-text pair, as there are normally unlimited number (up to the total number of images or texts in a data set) of negative image-text pairs, it is crucial to include a sufficiently large number of negative pairs in a contrastive loss to make the representation learning effective, as validated in all related works such as CLIP~\cite{clip_radford2021learning}, Florence~\cite{florence_yuan2021florence}, OpenCLIP~\cite{openclip_ilharco_gabriel_2021_5143773}, and BASIC~\cite{basic_pham2021combined}. Specifically, BASIC shows that larger batch size, plus larger data set and larger model, 
theoretically lead to a better generalization performance.

However, a fundamental technical challenge in training a CLIP-like model is how to enlarge its batch size under the constraint of limited GPU memory. For instance, when the batch size is 65,536, the similarity matrix for all image-text pairs in the batch will cost about 16GB using Float32. 
As the backbone part also consumes a significant portion of GPU memory, especially for large backbones such as ViT-Large or ViT-Huge~\cite{vit_dosovitskiy2020image}, scaling up batch size presents a great challenge, usually requiring hundreds of V100 or A100 GPUs~\cite{clip_radford2021learning, florence_yuan2021florence, openclip_ilharco_gabriel_2021_5143773}, which are inaccessible for most research scientists.

In this work, we develop a distributed solution called DisCo-CLIP for constrastive loss computation, which can save a large amount of memory for contrastive loss and make CLIP training more memory-efficient. Our method starts from a decomposition of the original contrastive loss. Based on this decomposition, we divide the contrastive loss into two parts, one to calculate the intra-GPU loss and gradients, and the other one to calculate the inter-GPU loss and gradients. For a mini-batch on the $n$-th GPU (hereinafter called its hosting GPU), its intra-GPU gradients are calculated on its hosting GPU, and its inter-GPU gradients are collected from other GPUs. DisCo is an exact solution, mathematically equivalent to the original non-distributed contrastive loss, but more memory- and computation-efficient.
It can decrease the memory cost of contrastive loss from $\bigO(B^2)$ to $\bigO(\frac{B^2}{N})$, where $B$ and $N$ are the batch size and the number of GPUs. When $N$ equals to 64, it means around 97\% (see Sec.~\ref{sec:Distri} for details) of the memory, and similarly the computational cost, in contrastive loss can be saved. Thus, using DisCo in CLIP, we can enable contrastive training with a larger batch size. Using 8 Nvidia A100 40GB GPUs, DisCo-CLIP can enable contrastive training of a ViT-B/32 model with a batch size of 32,768. Using 64 A100 40GB GPUs, DisCo-CLIP can train the same model with a larger batch size of 196K.

We summarize our contributions in twofold.
\begin{itemize}
    \item We propose a novel distributed contrastive loss solution called DisCo for memory efficient CLIP training, which can significantly reduce the memory consumption of the contrastive loss computation. Such a solution enables a larger batch size for contrastive training using the same computing resource without sacrificing any computation accuracy.
    \item We further validate that training with a larger batch size can further improve the performance of contrastive learning models.
\end{itemize}

\section{Background and Related Work}
\label{sec:related_works}
In this section, we will introduce some background information for contrastive language-image pre-training (CLIP) and review some works that reduce memory consumption of the backbone part by trading computation for memory.
\subsection{Contrastive Language-Image Pre-training}
\label{sec:rw_clip}
The idea behind CLIP~\cite{clip_radford2021learning} is to learn two representations via two encoders, an image encoder~\cite{resnet_he2016deep, bit_kolesnikov2020big, vit_dosovitskiy2020image, swint_liu2021swin, cvt_wu2021cvt} and a text encoder~\cite{bert_devlin2018bert, t5_raffel2020exploring, gpt3_brown2020language}.  Its target is to encourage positive image-text pairs to have higher similarities, and enforce negative image-text pairs to have lower similarities.
The training is supervised by two contrastive losses, an image-to-text loss and a text-to-image loss.
Suppose we have $B$ text-image pairs as a batch sending to two encoders and the model is trained using $N$ GPUs, with each GPU assigned $b=\frac{B}{N}$ pairs. Here, we use $\rT_{A}, \rI_{A}$ to denote all text and image features obtained from two encoders. Suppose the hidden dimension is $D$, then the shapes of $\rT_{A}, \rI_{A}$ are both $B\times D$. In this way, our contrastive losses can be written as,

\begin{equation} \label{eq:origianlclip}
\centering 
\cL = \cL_1(\rI_{A}, \rT_{A}) + \cL_2(\rT_{A}, \rI_{A}),
\end{equation}
where $\cL_1(\rI_{A}, \rT_{A})$ denotes the image-to-text loss, and $\cL_2(\rT_{A}, \rI_{A})$ represents the text-to-image loss. The image-to-text loss means that the loss is to encourage a given image to find
its paired text from tens of thousands of texts, and similarly for the text-to-image loss.
%mentioning the image-to-text loss, we mean, given an image, the loss is to find 

Motivated by the success of CLIP, many new methods have been proposed, such as FILIP~\cite{filip_yao2021filip}, LiT~\cite{lit_zhai2022lit}, ALIGN~\cite{align_jia2021scaling}, BASIC~\cite{basic_pham2021combined}, BLIP~\cite{blip_li2022blip},  GIT~\cite{git_wang2022git}, and K-LITE~\cite{klite_shen2022k}. FILIP attempts to obtain a fine-grained alignment between image patches and text words. In contrast, CLIP only obtains an image-level alignment between an image and a text sentence. FILIP achieves this goal by a modified contrastive loss. Instead of training both image and text models from scratch, LiT~\cite{lit_zhai2022lit} shows that employing a pre-trained image model and locking it would greatly benefit zero-shot classification. Under the contrastive framework, ALIGN~\cite{align_jia2021scaling} and BASIC~\cite{basic_pham2021combined} investigate how the scaling up of model, data, and batch size benefits contrastive learning. Instead of only leveraging contrastive learning framework, BLIP~\cite{blip_li2022blip} introduces a mixed contrastive learning and generative learning for vision-language model. Further, GIT~\cite{git_wang2022git} employs a pure generative learning framework and demonstrates its superior performance.

The success of CLIP and the above methods highly depends on  large-scale paired image-text data set~\cite{cc12_changpinyo2021conceptual, yfcc_thomee2016yfcc100m, wit_srinivasan2021wit, align_jia2021scaling, laion400_schuhmann2021laion, laion5b_schuhmann2022laion}. Compared with classification data sets, such as ImageNet 1K~\cite{imagenet_deng2009imagenet} and ImageNet 21K~\cite{imagenet21_ridnik2021imagenet}, which require careful human annotation, there are abundant paired image-text data on the Web and can be more easily collected~\cite{laion5b_schuhmann2022laion}. 

Large batch size is a prerequisite for vision-language contrastive learning. 
Data Parallelism (DP) and Model Parallelism (MP) are two widely used methods in deep learning for distributed training~\cite{sgd32k_you2017scaling, bert76_you2019large, transformer_vaswani2017attention, bert_devlin2018bert, onehour_goyal2017accurate, t5_raffel2020exploring}. There are many related works on this topic, such as DeepSpeed~\cite{zero_rajbhandari2020zero, zeroinfi_rajbhandari2021zero, deepspeed_rasley2020deepspeed} and Colossal-AI~\cite{colossal_bian2021colossal}. We recommend interested readers to refer to these papers for some common distributed learning techniques. In the following, we will describe several methods that trade computation for memory in the backbone part.

\subsection{Trade Computation for Memory in Backbone}
\label{sec:tradememory}

Checkpointing~\cite{checkpoint_chen2016training} is an effective approach to reduce memory consumption, especially for the intermediate results of low-cost operation. It only stores feature maps of some high-cost operations, such as convolution, MLP, and self-attention, but drops feature maps of low-cost operations such as activation (e.g., ReLU, GeLU) and layer normalization in the forward pass. In the backward process, these dropped feature maps can be recomputed quickly at a low cost. By dropping some intermediate feature maps, we can save a great amount of memory. Gruslys~\etal \cite{meb_gruslys2016memory} further extend the idea of Checkpointing to the recursive neural network. Checkpointing is helpful but still not sufficient if we want to further increase the batch size. 

Recently, BASIC~\cite{basic_pham2021combined} introduces a gradient accumulating (GradAccum) method for memory saving in backbone via micro-batching the contrastive loss. This approach is based on the observation that when computing the contrastive loss, we only need the full similarity matrix instead of all intermediate results. Thus, BASIC proposes to trade computation for memory by dropping some intermediate hidden states during the forward process, and then re-compute them during back-propagation.
This method can effectively reduce the memory consumption in the backbone part but it requires one extra forward computation. A similar gradient cache approach~\cite{scalingdcl_gao2021scaling} was also introduced for dense passage retriever in natural language processing.

Reversible model structure~\cite{revnet_gomez2017reversible,reformer_kitaev2019reformer} is another elegant method for memory saving in backbone. However, it requires block adaption because it needs to reconstruct the input layer according to the output layer in the back propagation process. Reversible structure is particularly suitable for models that do not change activation shape in the whole network.
However, reconstructing a layer from its output inevitably takes more backward time.

All above three mechanisms mainly focus on reducing memory consumption in the backbone part. However, none of them pays attention to reducing the memory consumption of the contrastive loss, which has been a big bottleneck for large batch size contrastive training. We will explain this problem in the next section.

\begin{figure}[t]
  \centering
   \includegraphics[width=1.0\linewidth]{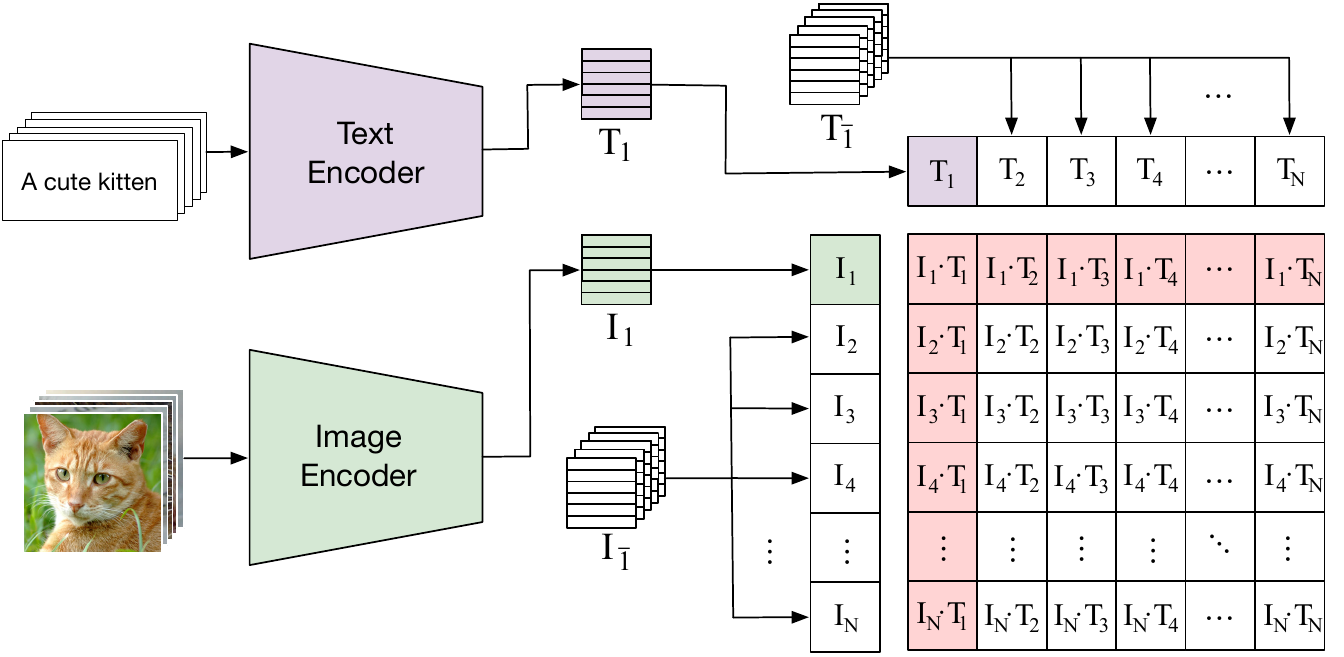}
   \caption{Illustration of the original CLIP training. All image features $\rI_A$ and text features $\rT_A$ are gathered to every GPU via \texttt{all\_gather} to calculate the $B\times B$ similarity matrix for contrastive loss computation. This computation is performed repeatedly on every GPU.}
   \label{fig:originalclip}
\end{figure}

%Let us see an example of data parallel (DP), in which activations in the network cost most of GPU memory. 

\section{Motivation}
\label{sec:Motivation}
Distributed contrastive learning is different from traditional distributed data parallel and model parallel strategies. 
Data parallel (DP) is particularly effective when activations in the network cost most of the GPU memory.
For example, in~\cite{onehour_goyal2017accurate}, Goyal~\etal succeeded in training an ImageNet classification model in one hour. Each GPU independently performs forward and backward computation. Only after backward computation, the GPUs perform a gradient communication to merge gradients. In contrast, model parallel (MP) is more effective when model parameters (\eg. 10 billions of parameters) consume a large mount of GPU memory. In this way, the model parameters are sliced and assigned to different GPUs. 
However, different from the DP and MP problems, in contrastive training, besides the backbone and model parameters, the contrastive loss consumes a larger amount of GPU memory especially when a large batch size is used.

%by slicing and assigning model parameters to .. In contrast, in model parallel (MP), model parameters usually consume a large mount of GPU memory. 

As we have discussed in the introduction section, a large batch size is of crucial importance for contrastive learning. Previous research~\cite{basic_pham2021combined} also proves theoretically that a larger contrastive batch size can lead to a better representation for image-text models. Interested readers can refer to \textit{Theorem 1} in~\cite{basic_pham2021combined} and its appendix for the proof.

However, a large batch in contrastive loss will consume a large amount of GPU memory, usually of $\bigO(B^2)$, where $B$ is the batch size. Let us take an examination of the memory consumption of contrastive loss. For example, for a batch 
size of 65,536, it stores a similarity matrix of 4.3 billions parameters which requires 16GB of memory using Float32. Its gradient computation in the back-propagation process will take another 16GB of memory. Considering other memory cost of model parameters and backbone intermediate activations, it cannot be trained using A100 40GB GPUs even with mixed precision computation.

After investigating the constrastive loss in depth, we notice that there is much redundant memory consumption and computational waste in the current CLIP training solution. As shown in Fig.~\ref{fig:originalclip}, in the original CLIP training, to calculate the gradients for the text and image features in the hosting GPU, every GPU in the distributed training environment needs to compute the entire $B\times B$ similarity matrix, which is not necessary as we will show later.

%which we will show is not necessary. 

Our motivation in this paper is to eliminate memory redundancy and remove unnecessary calculation via a loss decomposition, and thus enable large batch training using limited GPU resource.

\section{DisCo-CLIP: A Distributed Contrastive Loss for
Memory Efficient CLIP Training}
\label{sec:DisCoCLIP}
In this section, we introduce DisCo, a distributed solution for contrastive
loss computation. 
Using DisCo in CLIP, we accomplish a new memory efficient CLIP training method, called DisCo-CLIP. To this end, we decompose the loss calculation into two parts, one part to compute the intra-GPU loss and gradients and the other part to compute inter-GPU loss and gradients. This section will first introduce how we decompose this loss calculation. Then, we will describe the algorithm implementation of DisCo-CLIP in detail.

\subsection{DisCo: A Distributed Contrastive Loss}
\label{sec:Distri}
As shown in Eq.~\ref{eq:origianlclip}, $\rI_A$ and $\rT_A$ denote all image and text features collected from all GPUs. Here, we use $\rI_n$ and $\rT_n$ to denote the image and text features on the $n$-th GPU, and use $\rI_{\overline{n}}$ and $\rT_{\overline{n}}$ to denote the image and text features on all other GPUs. The division is denoted as,
\begin{equation} \label{eq:featuredecomp}
\centering
\begin{aligned}
\rI_{A} &= [\rI_n,\ \ \rI_{\overline{n}}], \\ 
\rT_{A} &= [\rT_n,\ \rT_{\overline{n}}]
\end{aligned}
\end{equation}
%where $\rI_n$ and $\rT_n$ denote the features computed in the $n$-th GPU, and $\rI_{\overline{n}}$ and $\rT_{\overline{n}}$ denote the features computed in all other GPUs. 
The shapes of $\rI_A$ and $\rT_A$ are $B\times D$, the shapes of $\rI_n$ and $\rT_n$ are $\frac{B}{N}\times D$, and the shapes of $\rI_{\overline{n}}$ and $\rT_{\overline{n}}$ are $\frac{B(N-1)}{N} \times D$.

%\small

%\begin{equation} 
%\centering 
%\cL = \cL_1(\rI_n, \rT_{A}) + \cL_1(\rI_{\on}, \rT_{A}) + \cL_2(\rT_n, \rI_{A}) + %\cL_2(\rT_{\on}, \rI_{A}),
%\end{equation}

According to the above definition, the contrastive loss can be decomposed and rewritten as,
\begin{equation}
\label{eq:cliplossfourparts}
\begin{aligned}
\cL_d = \cL_1(\rI_n, \rT_{A}) + \cL_1(\rI_{\on}, \rT_{A}) + \\
\cL_2(\rT_n, \rI_{A}) + \cL_2(\rT_{\on}, \rI_{A}),
\end{aligned}
\end{equation}
where $\cL_1(\rI_n, \rT_{A})$ denotes the image-to-text contrastive loss between image features $\rI_n$ and text features $\rT_A$, and $\cL_2(\rT_n, \rI_{A})$ denotes the text-to-image loss between $\rT_n$ and $\rI_A$. Mathematically, the computational result $\cL_d$ in Eq.~\ref{eq:cliplossfourparts} is the same as the result $\cL$ in  Eq.~\ref{eq:origianlclip}. However,
by decomposing the loss $\cL_d$ into four parts, the gradient flow in the back-propagation process is more obvious.
 Meanwhile, we can see that $\cL_1(\rI_{\on}, \rT_{A})$ does not induce gradients with respect to $\rI_n$ and $\cL_2(\rT_{\on}, \rI_{A})$ has no gradients  with respect to $\rT_n$. Actually, these two terms are redundant computation on the $n$-th GPU in the original contrastive loss, which unnecessarily consume a large amount of memory.

%we can easily see the gradient flow in the back-propagation process.

Thus, according to the decomposition in Eq.~\ref{eq:cliplossfourparts}, we can calculate the gradients $\frac{\partial \cL_d}{\partial \rI_n}$ for image features $\rI_n$, and the gradients $\frac{\partial \cL_d}{\partial \rT_n}$ for  text features $\rT_n$ as follows,
\begin{equation} \label{eq:lossdecomp}
\centering
\footnotesize
\begin{aligned}
\frac{\partial \cL_d}{\partial \rI_n}  &= \eqnmarkbox[blue]{Psi2}{\frac{\partial \cL_1(\rI_n, \rT_{A}) }{\partial \rI_n} + \frac{\partial \cL_2(\rT_n, \rI_{A}) }{\partial \rI_n}} + \eqnmarkbox[red]{Psi1}{\frac{\partial \cL_2(\rT_{\on}, \rI_{A}) }{\partial \rI_n}}, \\
\frac{\partial \cL_d}{\partial \rT_n}  &= \eqnmarkbox[blue]{Psi2}{\frac{\partial \cL_1(\rI_n, \rT_{A}) }{\partial \rT_n} + \frac{\partial \cL_2(\rT_n, \rI_{A}) }{\partial \rT_n}} + \eqnmarkbox[red]{Psi1}{\frac{\partial \cL_1(\rI_{\on}, \rT_{A}) }{\partial \rT_n}},
\end{aligned}
\end{equation}
where the losses in the red part can be further unfolded in a sum form as,
\begin{equation} \label{eq:negativelosses}
\footnotesize
\centering 
\begin{aligned}
\frac{\partial \cL_2(\rT_{\on}, \rI_{A}) }{\partial \rI_n} &= \sum_{i\neq n \And i \in [1,N]} \frac{\partial \cL_2(\rT_{i}, \rI_{A}) }{\partial \rI_n}, \\
\frac{\partial \cL_1(\rI_{\on}, \rT_{A}) }{\partial \rT_n} &= \sum_{j\neq n \And j \in [1,N]} \frac{\partial \cL_1(\rI_{j}, \rT_{A}) }{\partial \rT_n}.
\end{aligned}
\end{equation}

The gradients of $\frac{\partial \cL_d}{\partial \rI_n}$ and $\frac{\partial \cL_d}{\partial \rT_n}$ both consist of three terms. In both equations, all three terms can be divided into two parts, we mark them with two different colors, \textcolor{blue}{blue} and \textcolor{red}{red}.
The gradients in the \textcolor{blue}{blue} part are calculated on the $n$-th (hosting) GPU, and the gradients in the \textcolor{red}{red} part are calculated on other GPUs. We call them intra-GPU gradients and inter-GPU gradients, respectively. When computing the \textcolor{red}{red} part for $\frac{\partial \cL_d}{\partial \rI_n}$, the images $\rI_n$ are considered as negative samples. It should be noted that although $\rI_n$ is regarded as negative samples, there still induce gradients for $\rI_n$. 

%still exist gradients for

\begin{figure}[t]
  \centering
   \includegraphics[width=0.98\linewidth]{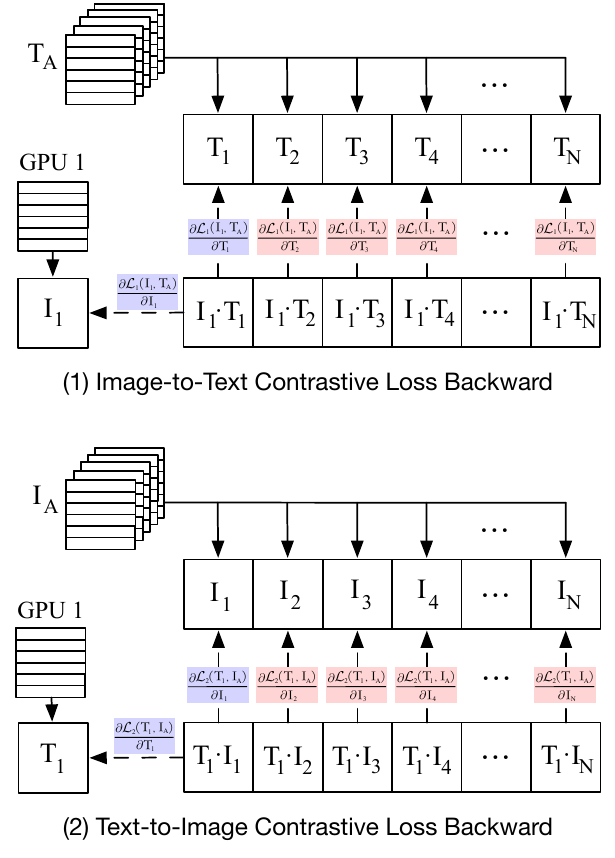}
 \caption{Illustration of the gradient calculation of DisCo on the first GPU. In DisCo, to calculate the gradients on the current GPU, we only need to compute two $\frac{B}{N}\times B$ similarity matrices instead of a full $B\times B$ matrix as shown in Fig.~\ref{eq:origianlclip}.}
\label{fig:discoclip1}
\end{figure}

According to this decomposition, to compute $\frac{\partial \cL_d}{\partial \rI_n}$ and $\frac{\partial \cL_d}{\partial \rT_n}$, the $n$-th GPU only needs to compute the following four terms,
\begin{equation}\label{eq:fourterms}
\footnotesize
  \eqnmarkbox[blue]{Psi2}{\left[\frac{\partial \cL_1(\rI_n, \rT_{A}) }{\partial \rI_n}, \frac{\partial \cL_2(\rT_n, \rI_{A}) }{\partial \rI_n}, \frac{\partial \cL_1(\rI_n, \rT_{A}) }{\partial \rT_n}, \frac{\partial \cL_2(\rT_n, \rI_{A}) }{\partial \rT_n}\right]}.
\end{equation}
All four terms only need to compute two small similarity matrices of shape $\frac{B}{N}\times B$ instead of a full matrix of shape $B\times B$ which consumes a large amount of memory. In this way, we can reduce memory consumption from $B^2$ to $\frac{2B^2}{N}$. For instance, when $N=16$, we save $\frac{7}{8}$ of memory consumption, if $N$ is 64, we save $\frac{31}{32}$ of memory consumption.
Meanwhile, it also saves much computation. The computational cost for the contrastive similarity matrix decreases from $B^2D$ to $\frac{2B^2 D}{N}$.

For the two red terms in Eq.~\ref{eq:lossdecomp}, 
\begin{equation}\label{eq:redtwoterms}
\footnotesize
\eqnmarkbox[red]{Psi1}{\left[\frac{\partial \cL_2(\rT_{\on}, \rI_{A}) }{\partial \rI_n}, \frac{\partial \cL_1(\rI_{\on}, \rT_{A}) }{\partial \rT_n}\right]},
\end{equation}
we use the \texttt{all\_reduce} operation to perform gradient communication and collect all gradients with respect to $\rI_n$ and $\rT_n$. Fig.~\ref{fig:discoclip1} illustrates the gradient calculation of DisCo on the first of $N$ GPUs.

Let us compare DisCo with the original contrastive loss. In the original implementation, because of the lack of gradient communication in the process of back-propagation among GPUs\footnote{Gradient communication here in the process of back-propagation should be distinguished from the gradient collection after model back-propagation which is to collect and average the weight gradients.}, each GPU must compute the whole similarity matrix between $\rI_A$ and $\rT_A$ to obtain the gradients $\frac{\partial \cL_d}{\partial \rI_n}$ and $\frac{\partial \cL_d}{\partial \rT_n}$ on the $n$-th GPU. In DisCo, for each GPU, we only need to calculate a subset instead of the whole similarity matrix. We then use an \texttt{all\_reduce} operation to conduct gradient communication and collect the gradients.
The distributed contrastive loss can also be applied to the contrastive loss computation in self-supervised learning~\cite{mocov3_chen2021empirical, simclr_chen2020simple}.

\subsection{Algorithm Implementation of DisCo-CLIP}
\label{sec:implementation}
\begin{algorithm}[t]
\scriptsize %\small, \footnotesize, \scriptsize, or \tiny
\SetAlgoLined
    \PyComment{image\_encoder: ResNet or ViT} \\
    \PyComment{text\_encoder: text Transformer} \\
    \PyComment{img[b,H,W,C]: minibatch of images} \\
    \PyComment{text[b,L]: minibatch of texts} \\
    \PyComment{t: temperature parameter} \\
    \PyComment{D: feature dimension} \\
    \PyComment{N: nums of GPU} \\
    \PyComment{rank: global rank} \\
    %\PyComment{th: torch} \\
    
    \
    
    \PyComment{extract feature representations} \\
    \PyCode{i\_e = image\_encoder(img)} \PyComment{[b,D]}\\
    \PyCode{t\_e = text\_encoder(text)} \PyComment{[b,D]} \\
    
    \
    
    \PyComment{gather features from all gpus} \\
    \PyCode{I\_E = all\_gather(i\_e)} \PyComment{[N,b,D]} \\
    \PyCode{T\_E = all\_gather(t\_e)} \PyComment{[N,b,D]} \\
    
    \ 
    
    \PyComment{scaled dot product similarities} \\
    \PyCode{logits\_i = dot(I\_E[rank], T\_E)*t} \PyComment{[b,N*b]} \\
    \PyCode{logits\_t = dot(T\_E[rank], I\_E)*t} \PyComment{[b,N*b]} \\
    
    \
    
    \PyComment{image-to-text and text-to-image losses} \\
    \PyCode{labels = arange(b) + b * rank} \\
    \PyCode{loss\_i = cross\_entropy\_loss(logits\_i, label)} \\
    \PyCode{loss\_t = cross\_entropy\_loss(logits\_t, label)} \\
    
    \
    
    \PyComment{loss backward } \\
    \PyCode{loss = (loss\_i + loss\_t) / 2} \\
    \PyCode{loss.backward()} \\
    
    \ 
    
    \PyComment{reduce gradients and losses from all gpus} \\
    \PyCode{I\_E.grad = all\_reduce(I\_E.grad, op=AVG)} \\
    \PyCode{T\_E.grad = all\_reduce(T\_E.grad, op=AVG)} \\
    \PyCode{Loss = all\_reduce(loss, op=AVG)} \\
    \ 
    
    \PyComment{backbone backward } \\
    \PyCode{i\_e.backward(I\_E.grad[rank])} \\
    \PyCode{t\_e.backward(T\_E.grad[rank])} \\
    
\caption{Pseudo-code for DisCo-CLIP}
\label{algo:your-algo}
\end{algorithm}

%$\frac{\partial \cL_d}{\partial \rI_n}$ and $\frac{\partial \cL_d}{\partial \rT_n}$, 
%$\frac{\partial \cL_d}{\partial \rI_n}$ and $\frac{\partial \cL_d}{\partial \rT_n}$

Using DisCo in CLIP, we develop a memory efficient CLIP training method, termed as DisCo-CLIP.
To make our algorithm clearer, we show the pseudo-code for DisCo-CLIP in Alg.~\ref{algo:your-algo}. The implementation can be roughly divided into three segments. In the first segment, for each GPU with a global rank number, we extract its image and text features, and then use \texttt{all\_gather} to collect image and text features into this GPU. Then in the second segment, we compute the similarity matrix and the two losses: image-to-text and text-to-image, and use the losses to compute the gradients. In this segment, 
the intra-GPU gradients, 
$\left[\frac{\partial \cL_1(\rI_n, \rT_{A}) }{\partial \rI_n}, \frac{\partial \cL_2(\rT_n, \rI_{A}) }{\partial \rI_n}, \frac{\partial \cL_1(\rI_n, \rT_{A}) }{\partial \rT_n}, \frac{\partial \cL_2(\rT_n, \rI_{A}) }{\partial \rT_n}\right]$, 
as defined in Eq.~\ref{eq:fourterms}, will be computed on the current GPU. In the last segment, we use \texttt{all\_reduce} to finish gradient communication. In this segment, the inter-GPU gradients, $\left[\frac{\partial \cL_2(\rT_{\on}, \rI_{A}) }{\partial \rI_n}, \frac{\partial \cL_1(\rI_{\on}, \rT_{A}) }{\partial \rT_n}\right]$,  as defined in Eq.~\ref{eq:redtwoterms}, will be collected to the current GPU. The intra-GPU and inter-GPU gradients are averaged to obtain $\left[\frac{\partial \cL_d}{\partial \rI_n}, \frac{\partial \cL_d}{\partial \rT_n}\right]$. Finally, we perform backward propagation for the backbone.

DisCo-CLIP has two advantages compared with CLIP. First, it decreases the memory consumption for the contrastive loss from $\bigO(B^2)$ to $\bigO(\frac{B^2}{N})$. Second, it reduces the computational complexity of the contrastive loss from $\bigO(B^2D)$ to $\bigO(\frac{B^2}{N}D)$. Besides these two advantages, DisCo-CLIP is an exact solution for CLIP without sacrificing any computation accuracy.

Note that DisCo-CLIP requires an extra \texttt{all\_reduce} operator to perform gradient communication, which has the same cost as the \texttt{all\_gather} operator. Compared with the full similarity matrix computation, this extra communication cost is quite small and negligible. See Table~\ref{tab:speedevaluation} for a speed comparison.

For the backbone part, we can also use the GradAccum strategy as proposed in BASIC~\cite{basic_pham2021combined}. 
BASIC provides a simple yet effective memory-efficient strategy for backbone memory optimization.
Combined with GradAccum, DisCo-CLIP can further lead to a more memory-efficient solution. 

\begin{table}[ht]
\footnotesize
  \centering
  \begin{tabular}{l@{\ \ \ }l@{\ \ \ }l@{\ \ \ }l}
  \hline 
  Methods & Backbone & Contra. Loss & Overall \\ 
  \hline 
  CLIP\cite{clip_radford2021learning} & \(\bigO\left( \frac{B}{N}LD \right)\) & \(\bigO\left(B^2 \right)\) & \(\bigO(\frac{B}{N}LD)\) + \(\bigO(B^2)\) \\ 
  BASIC\cite{basic_pham2021combined} & \(\bigO\left(\frac{B}{N}D \right)\) & \(\bigO\left(B^2  \right)\)  &  \(\bigO(\frac{B}{N}D)\)+\(\bigO(B^2)\)  \\ 
  DisCo-CLIP & \(\bigO\left(\frac{B}{N}LD \right)\) & \(\bigO\left(\frac{B^2}{N} \right)\) & \(\bigO(\frac{B}{N}LD)\) + \(\bigO(\frac{B^2}{N})\)  \\ 
  $\text{DisCo-CLIP}^{*}$ & \(\bigO\left(\frac{B}{N}D \right)\) & \(\bigO\left(\frac{B^2}{N} \right)\) & \(\bigO(\frac{B}{N}D)\) + \(\bigO(\frac{B^2}{N})\)  \\ 
  \hline
  \end{tabular}
  \caption{Memory consumption comparison of different methods. $\text{DisCo-CLIP}^{*}$ denotes the combination of DisCo-CLIP and the GradAccum strategy in BASIC. $B$ denotes the batch size, which is usually very large, e.g. 32K, $N$ is the number of GPUs, $L$ is the number of layers in the backbone, and $D$ is the feature dimension. For simplicity, we use the same $D$ for all layers.}
  \label{tab:memorycomparison}
\end{table}

A detailed comparison of memory consumption for CLIP, BASIC, and DisCo-CLIP is provided in Table~\ref{tab:memorycomparison}. With GradAccum, BASIC can reduce the overall memory consumption from $\bigO(\frac{B}{N}LD) + \bigO(B^2)$ to $\bigO(\frac{B}{N}D) + \bigO(B^2)$. DisCo-CLIP can reduce the memory consumption to $\bigO(\frac{B}{N}LD) + \bigO(\frac{B^2}{N})$, $\text{DisCo-CLIP}^{*}$ can further reduce the memory consumption to $\bigO(\frac{B}{N}D) + \bigO(\frac{B^2}{N})$. 
A practical example is that, in vanilla CLIP~\cite{clip_radford2021learning} based on ViT-B/32, $D=1024, B=32768, N=128$.

\section{Experiments}
\label{sec:experiments}
In this section, we first describe the experimental settings including data sets and pre-training details. Then, we present the experiments regarding memory consumption, training efficiency, batch size, and data scale. Finally, we report zero-shot classification on several data sets.
\subsection{Experimental Setting}
\textbf{Image-Text Data Sets.} A large-scale image-text data set is a prerequisite for contrastive image-text pre-training. In the literature, there are many private~\cite{wit_srinivasan2021wit, align_jia2021scaling} or public data sets~\cite{yfcc_thomee2016yfcc100m, laion400_schuhmann2021laion, laion5b_schuhmann2022laion}.  LAION-400M~\cite{laion400_schuhmann2021laion} is a data set with 400 million image-text pairs filtered by CLIP score. We crawl the images according to the provided URLs and finally obtain 360M image-text pairs as some URLs are expired and no longer available. LAION-5B~\cite{laion5b_schuhmann2022laion} is a larger data set with 5 billion image-text pairs. %The LAION group~\footnote{\url{https://laion.ai/blog/laion-5b/}} also provides the URLs and other information. 
It contains a subset of 2.32 billion English image-text pairs. We refer to this subset as LAION-2B as in~\cite{laion5b_schuhmann2022laion}. We also crawl the data set according to the released URLs and finally obtain 2.1 billion image-text pairs. 
In this paper, for a fair comparison with CLIP~\cite{clip_radford2021learning} and LAION~\cite{laion5b_schuhmann2022laion} which conduct experiments using 400M data pairs, we create a subset of 400M image-text pairs from our crawled LAION-2B and also make a subset of 100M image-text pairs for ablation study.

\textbf{Pre-training Details.} The implementation of our DisCo-CLIP is based on OpenCLIP\footnote{\url{https://github.com/mlfoundations/open_clip}}~\cite{openclip_ilharco_gabriel_2021_5143773}, an open source CLIP implementation. We follow the original CLIP\footnote{\url{https://github.com/openai/CLIP}}~\cite{clip_radford2021learning} and mainly use ViT-B/32~\cite{vit_dosovitskiy2020image} as the vision backbone.% although some other neural networks~\cite{resnet_he2016deep, bit_kolesnikov2020big, swint_liu2021swin, cvt_wu2021cvt} are also optional. 
Training images are resized to $224\times 224$ and fed to the vision backbone. For the text backbone, we use a text Transformer~\cite{transformer_vaswani2017attention, bert_devlin2018bert}, and set the maximum length of text to 76 as in ~\cite{clip_radford2021learning}. Due to limited time and GPU resource, we do not train ViT-L/14 because it takes more than four times of training time compared to ViT-B/32. It should also be noted that in all our experiments, we do not use the GradAccum strategy as in BASIC~\cite{basic_pham2021combined}. Checkpointing is enabled in vanilla CLIP and our DisCo-CLIP.
We train models using an Adam~\cite{adam_kingma2014adam} optimizer and cosine learning rate scheduler with a linear warmup~\cite{warmup_loshchilov2016sgdr}. Weight decay regularization~\cite{adamw_loshchilov2018decoupled} is used on all parameters except bias, layer normalization, and temperature in contrastive loss. Since our training experiments use different batch sizes, according to previous works~\cite{onehour_goyal2017accurate, t5_raffel2020exploring}, we change the learning rate for different batch sizes. For our learning rate scheduler, we first define a base learning rate of 5e-4 and a base batch size of 32,768, and then linearly warm it up to the maximum learning rate. The maximum learning rate is determined according to  a linear strategy, $\text{max\_lr = base\_lr} * \frac{\text{batch size}}{\text{32,768}}$. Training stability is a challenging issue in mixed-precision training as NaN might happen in many cases.
We obverse that $\beta_2$ in Adam is very important for training stability and find 0.98 works well for our experiments instead of the default value of 0.999. 
For most of our experiments, we use 64 NVIDIA A100 40GB GPUs for model training, but for some experiments, we only use 8 NVIDIA A100 40GB GPUs to show that our solution is more resource-efficient. 

%\textbf{DisCo-CLIP $\equiv$ CLIP under the same setting.} 
\subsection{Evaluation Results}
\textbf{Equivalence Validation.} To empirically validate our 
theoretical analysis, we conduct experiments to compare CLIP and DisCo-CLIP. We use the same random seed to initiate the network parameters and use the same batch size 32,768 on 64 GPUs, and train ViT-B/32 models with 16 epochs on LAION-100M. %The results are reported in Table~\ref{tab:clipreproduction}.

\begin{table}[ht]
\small
    \centering
    \begin{tabular}{cccc}
    \hline 
    Methods & Memory & Accuracy  & Time (Hours) \\ 
    \hline 
     CLIP      & {27.4GB}  & 51.64 & 13.4  \\ 
     DisCo-CLIP   & {16.5GB}  & 51.64 & 12.1  \\
    \hline
    
    \end{tabular}
    \caption{CLIP Reproduction. Under the same training settings, DisCo-CLIP has exactly the same accuracy (top-1 zero-shot classification accuracy on ImageNet) as CLIP, but uses lower memory consumption and less training time. Training resource: 64 A100 40GB GPUs.}
    \label{tab:clipreproduction}
\end{table}

From Table~\ref{tab:clipreproduction}, we see that DisCo-CLIP has exactly the same accuracy as CLIP, but uses lower memory consumption and less training time.
Under the same training settings, DisCo-CLIP and CLIP also have exactly the same training curves. This experiment further validates the numerical equivalence between DisCo-CLIP and CLIP.

%and the same zero-shot classification on ImageNet. Essentially, DisCo-CLIP just decreases memory consumption and remove redundant computation, and thus enable model training with larger batch size, but not affect the final loss computation of CLIP under the same batch size.

%DisCo-CLIP saves more computational time in loss computation. With larger batch size, DisCo-CLIP can save more computational time.

%Our DisCo-CLIP focuses on memory reduce on the contrastive loss, we also incorporate the BASIC, a method to decrease the backbone memory consumption, into our algorithm, we call the version as  $\text{DisCo-CLIP}^{*}$. For BASIC, we will use 16 chunks.
%Considering the parameters of our used networks are not too large, we will only need to evaluate the memory consumption of the backbone and the contrastive loss. 

\begin{table}[ht]
\small
    \centering
    \begin{tabular}{cccc}
    \hline 
    Methods & GPUs & BS & Memory \\ 
    \hline 
     CLIP        &  64$\times$ A100 40GB & 32,768  & {27.4GB}\\ 
     CLIP        &  64$\times$ A100 40GB & 65,536  & OOM \\ 
     DisCo-CLIP  &  64$\times$ A100 40GB & 32,768  & {16.5GB}\\ 
     DisCo-CLIP  &  64$\times$ A100 40GB & \textbf{196,608} & {38.1GB}\\ 
     DisCo-CLIP  &  \textbf{8}$\times$ A100 40GB & 32,768   & 36.9GB\\ 
    \hline
    
    \end{tabular}
    \caption{Memory consumption of CLIP and DisCo-CLIP under different settings. ``OOM'' means out of memory. }
    \label{tab:memoryconsumption}
\end{table}

\begin{table*}[h]
\footnotesize
    \centering
    \begin{tabular}{ccccc@{\ \ \ }c@{\ \ \ }c@{\ \ \ \ }cc@{\ \ }cc}
    \hline 
    \multirow{2}{*}{Methods} & \multirow{2}{*}{GPUs} & \multirow{2}{*}{Batch size} & \multirow{2}{*}{Backbone} &
    \multicolumn{3}{c}{Loss Operators} & \multirow{2}{*}{Loss Total} & \multicolumn{2}{c}{Backward} & \multirow{2}{*}{Total}  \\ \cmidrule(lr){5-7}  \cmidrule(lr){9-10}
    & &  & & all\_gather & Loss & all\_reduce & &Loss BP & Backbone BP &  \\
    \hline
    CLIP & \multirow{1}{*}{64} & \multirow{1}{*}{32,768} & \multirow{2}{*}{129.4}  &  {6.1} & {55.4}  & - &  {61.5}  & \multicolumn{2}{c}{414.8} & {605.6}\\
    
    DisCo-CLIP & \multirow{1}{*}{64} & \multirow{1}{*}{32,768} & &  {6.1} & {1.6}  & {4.3} & {\textbf{12.0}} $(80.5\%\downarrow)$  & {2.8} & {337.3} & {\textbf{481.5}} $(20.5\%\downarrow)$ \\
    \hline
    \end{tabular}
    \caption{Detailed comparisons of training efficiency of CLIP and DisCo-CLIP. All time-related measurements (the last nine columns) are in millisecond (ms). DisCo-CLIP saves more computational time in loss computation. Training resource: 64 A100 40GB GPUs.}
    \label{tab:speedevaluation}
\end{table*}

\textbf{Memory Consumption.} Besides memory consumption of network parameters, backbone and contrastive loss consumes most memory. We evaluate the memory consumption of the original CLIP and our DisCo-CLIP under different settings.  We report the peak memory consumption for different settings, and the results are shown in Table~\ref{tab:memoryconsumption}.

We have three observations from Table~\ref{tab:memoryconsumption}. 1) CLIP with ViT-B/32 can be trained with batch size 32,768 but fails with batch size 65,536. 
2) Our DisCo-CLIP can enable training of ViT-B/32 with a much larger batch size, 196,608 with 64 A100 40GB GPU. 3) DisCo-CLIP can enable training of the same model on a cluster with only 8 GPUs.

\textbf{Training Speed.} One iteration of CLIP consists of \textit{extracting features from backbone, gathering features from all GPUs, calculating loss, and executing back-propagation.} (We exclude some other factors, including data loading and optimizer update that are not directly related to the model computation or can be ignored.) Instead, DisCo-CLIP decomposes the back-propagation into two BP processes (Loss BP and Backbone BP) and have an additional \texttt{all\_reduce} operator. We evaluate the running time of each operator in CLIP and our DisCo-CLIP, and the results are shown in Table~\ref{tab:speedevaluation}.

From Table~\ref{tab:speedevaluation}, we can see that 1) for total loss computation, under the same batch size 32,768, we decrease the computation time from 61.5 ms to 12.0 ms. It is because in DisCo-CLIP, we only need to calculate two subset similarity matrices  between multiplying a  $\frac{B}{N}\times D$ matrix and a $B\times D$ matrix instead of a full matrix multiplication between two $B\times D$ matrices. 2) the operators of both \texttt{all\_reduce} and \texttt{all\_gather} are faster compared to the loss computation.
3) under the same batch size 32,768, DisCo-CLIP improves around $20.5\%$ total training efficiency. 
%Since DisCo-CLIP can enable larger batch training, and larger batch usually has better parallel efficiency, thus when training with larger batch size, the training efficiency of DisCo-CLIP further improves.

Overall, with DisCo-CLIP, we are able to not only train contrastive learning models with a larger batch size, but also improve the training speed considerably. DisCo-CLIP is a free-lunch solution for large-scale contrastive learning.

\begin{table}[ht]
\small
    \centering
    \begin{tabular}{cccc}
    \hline 
    Batch Size & Steps & Epochs & ViT-B/32\\ 
    \hline 
    8,192 & \(\approx 200 \mathrm{~K}\)  & 16 & \(48.76\) \\ 
    16,384 & \(\approx 100 \mathrm{~K}\) & 16 & \(50.95\) \\ 
    32,768 & \(\approx 50 \mathrm{~K}\) & 16 & \(51.64\) \\ 
    65,536 & \(\approx 25 \mathrm{~K}\)  & 16 & \(\mathbf{51.91}\)  \\ 
    % 131,072 & \(25 \mathrm{~K}\) & 16 & \(\mathbf{-}\)  \\ 
    \hline
    
    \end{tabular}
    \caption{Evaluation of different batch sizes on DisCo-CLIP.
    All models are trained for 16 epochs on a LAION-100M data set,  we show top-1 zero-shot classification accuracy on ImageNet. Training resource: 64 A100 40GB GPUs.}
    \label{tab:batchsize}
\end{table}

\textbf{Batch Size.} To assess the role of a larger batch size to the final performance, we conduct some controlled experiments for ViT-B/32 on our LAION-100M and LAION-400M subsets. All training hyper-parameters are the same except that we vary the batch size and the number of training steps.
To ensure all models ``see'' the same number of training samples, we use fewer training steps when training with a larger batch size. In Table~\ref{tab:batchsize}, we report the results of all settings.

From Table~\ref{tab:batchsize}, we can see that large batch size generally helps contrastive learning. The performance of using batch size 65,536 improves that of using batch size 32,768 by around 0.3\%. In the evaluation of data scaling, we further observe that the improvement widens when we train the model for more epochs.

\begin{table}[ht]
\small
    \centering
    \begin{tabular}{cccc}
    \hline 
    Batch Size  & 100M$\times$16 & 400M$\times$4 & 400M$\times$8 \\ 
    \hline 
    16,384  & \(50.95\) & \(54.36\) & \(58.38\) \\ 
    32,768  & \(51.64\) & \(55.52\) & \(59.69\) \\
    65,536  & \(51.91\) & \(55.50\) & \(\mathbf{59.96}\) \\ 
    \hline
    
    \end{tabular}
    \caption{Evaluation of data scaling on DisCo-CLIP. Three batch sizes are evaluated with two LAION subsets: LAION-100M and LAION-400M. The numbers $\times 16$, $\times 4$, and $\times 8$ are the training epochs for each data set. Training resource: 64 A100 40GB GPUs.}
    \label{tab:datascale}
\end{table}

\begin{table*}
\small
    \centering
    \begin{tabular}{ccccccccc}
    \hline 
    Model & Data Sets  & Epochs & Steps & Batch Size & INet~\cite{imagenet_deng2009imagenet}  & INet-v2~\cite{imagenetv2_recht2019imagenet}  & INet-R~\cite{imagenetr_hendrycks2021many}  & INet-S~\cite{imagenets_wang2019learning}  \\ 
    \hline 
    CLIP~\cite{clip_radford2021learning}       & CLIP WIT 400M & 32 & \(\approx {400} \mathrm{~K}\) & 32,768 & \(63.3\) & \(56.0\) & 69.4 & \(42.3\) \\ 
    OpenCLIP~\cite{openclip_ilharco_gabriel_2021_5143773,laion5b_schuhmann2022laion} & LAION-400M &32 & \(\approx {400} \mathrm{~K}\)  & 32,768 & \(62.9\) & \(55.1\) & 73.4 & \(49.4\)  \\ 
    DisCo-CLIP & LAION-400M$^{*}$ &32 & \(\approx {400} \mathrm{~K}\)  & 32,768 & \(63.2\) & \(55.2\) & 73.4 & \(50.6\)   \\ 
    DisCo-CLIP & LAION-400M$^{*}$ &32 & \(\approx {\textbf{200}} \mathrm{~K}\)  & \textbf{65,536} & \({\textbf{64.3}}\) & \(\textbf{56.2}\) & \textbf{73.8} & \(\textbf{51.7}\)  \\ 
    \hline
    \end{tabular}
    \caption{Comparison of DisCo-CLIP with vanilla CLIP and re-implemented CLIP by LAION group. We report top-1 zero-shot classification accuracy (\%)  on several data sets. All models are based on ViT-B/32. Our LAION-400M$^{*}$ is a 400M subset of LAION-2B. Training resource: 64 A100 40GB GPUs.}
    \label{tab:finalzeroshot}
\end{table*}
\textbf{Data Scaling.} To investigate the role of data scaling, we also conduct some controlled experiments for ViT-B/32 on the LAION 100M and LAION 400M subsets. For all models, we use three different batch sizes, 16,384, 32,768 and 65,536 for training. Models are trained with three different settings, 100M with 16 epochs, 400M with 4 epochs and 400M with 8 epochs.
Table~\ref{tab:datascale} reports the final ImageNet top-1 zero-shot classification accuracy of all models.

From Table~\ref{tab:datascale}, we observe that, ``seeing'' the same number of training samples (LAION-100M with 16 epochs, LAION-400M with 4 epochs.), the model trained on larger scale of data set can obtain better performance. For example, trained with the same batch size 65,536, the performance improves from 51.91\% to 55.50\% when the data set increases from 100M to 400M.  This observation validates the value of data scaling. 
This observation is also consistent with the combined scaling law in BASIC~\cite{basic_pham2021combined}. We also find that training with more epochs on larger scale of data with larger batch size will bring in larger performance gain. The gap between models (with batch sizes 32,768 and 65,536) trained on 400M with 4 epochs is around 0\% (from 55.52\% to 55.50\%), but it increases to 0.3\% (from 59.69\% to 59.96\%) when the models are trained for 8 epochs.
Thus, we can expect that if we use larger scale of data set, we should also use larger batch size to get a better performance.

\subsection{Zero-Shot Classification}
Following CLIP~\cite{clip_radford2021learning} and LAION-5B~\cite{laion5b_schuhmann2022laion}, we evaluate zero-shot top-1 classification accuracy of our models on several data sets, including INet~\cite{imagenet_deng2009imagenet}, INet-v2~\cite{imagenetv2_recht2019imagenet}, INet-R~\cite{imagenetr_hendrycks2021many} and INet-S~\cite{imagenets_wang2019learning}. We use exactly the same 80 prompts\footnote{\url{https://github.com/openai/CLIP/blob/main/notebooks/Prompt_Engineering_for_ImageNet.ipynb}} as CLIP.
Some standard prompts are like ``a bad photo of a \{\}.'', ``a photo of many \{\}.'', and etc. Following LAION-5B, we extract the embedding for each prompt using the text encoder (text Transformer), and average embeddings of all prompts in each class, and finally get the final class embedding. Given an image, we extract its image embedding via the image encoder (ViT-B/32), and compute its similarities with all class embeddings, and classify it into the class with the highest similarity score. %The results are shown in Table~\ref{tab:finalzeroshot}.

We can see  from Table~\ref{tab:finalzeroshot}, that using the same batch size of 32,768 as the OpenCLIP implementation, our DisCo-CLIP obtains a similar performance with OpenCLIP, the result of OpenCLIP is reported by LAION group in~\cite{laion5b_schuhmann2022laion}. Using a larger batch size of 65,536, DisCo-CLIP achieves a better performance than the OpenCLIP implementation and the original CLIP. Specifically, trained with the same 32 epochs, DisCo-CLIP using a batch size of 65,536 improves the OpenCLIP's result that is based on batch size of 32,768 by 1.4\%. Note that when doubling the batch size, we halve the training 
steps. This observation further validates that large batch size is important for contrastive learning.

%The results in Table~\ref{tab:finalzeroshot} further validates our evaluation in ablation study. Batch size matters, data scale matters.
%It should be noted that doubling the batch size, we will halve the training steps. 

\section{Discussion}
Recent developments in large-scale contrastive or generative learning, such as CLIP~\cite{clip_radford2021learning}, ALIGN~\cite{align_jia2021scaling}, FILIP~\cite{filip_yao2021filip}, LiT~\cite{lit_zhai2022lit}, BASIC~\cite{basic_pham2021combined}, LAION-5B~\cite{laion5b_schuhmann2022laion}, and GIT~\cite{git_wang2022git} have potential for many real-world applications, such as image classification, text-to-image retrieval, recommendation system, and generation models. Besides large-scale data sets, the progress highly depends on powerful GPU or TPU computing resources. As shown in Table~\ref{tab:hardware}, all methods use a large amount of computing hardwares. FILIP uses less GPU but also uses smaller batch size.
%For instance, CLIP uses 256 NVIDIA V100 GPUs, FILIP use 128 NVIDIA V100 GPUs, Lit uses 128 TPUs, LAION-5B uses 400 A100 GPUs, and BASIC uses a cluster of 2,048 TPUv3 cores.

\begin{table}[ht]
\small
    \centering
    \begin{tabular}{l l}
    \hline 
    Methods   & Hardware  \\ 
    \hline 
     CLIP~\cite{clip_radford2021learning}     & 256 NVIDIA V100 GPUs  \\ 
     FILIP~\cite{filip_yao2021filip}    & 128 NVIDIA V100 GPUs  \\
     LAION-5B~\cite{laion5b_schuhmann2022laion} & 128-400 NVIDIA A100 GPUs \\
     LiT~\cite{lit_zhai2022lit}      & 128-400 TPUs \\
     BASIC~\cite{basic_pham2021combined}    & 2,048 TPUs v3 \\
    \hline
    \end{tabular}
    \caption{Several vision-language contrastive learning methods and their used hardwares.}
    \label{tab:hardware}
\end{table}

Unfortunately, only a few big technology companies can afford this task. Large batch contrastive loss consumes a large amount of GPU memory, while large batch size is a prerequisite for contrastive learning. As a result, it is challenging for companies with limited resources or academic research institutes to 
contribute to research on this topic.
Our work provides a simple yet effective solution for reducing memory consumption that can help bridge the gap between academia and industry on this topic.

%small companies or academic research institutes cannot conduct research on this topic. 

%However, besides of memory requirement, we still observe that the large-scale data sets always need powerful distributed training. Thus, the gap stills exists.

%We expect we can build some standard, moderate scale of benchmark for evaluation in future, so more and more academic research institutes or small companies can contribute to this area.

\section{Conclusion}
In this paper, we have proposed a simple yet effective distributed contrastive loss solution called DisCo to reduce the memory consumption in contrastive learning. By decomposing the loss computation into intra-GPU computation and inter-GPU computation, we can remove redundant memory consumption and use a {\texttt{all\_reduce}} operator to collect inter-GPU gradients. Using DisCo in CLIP, the introduced DisCo-CLIP can enable a much larger batch contrastive learning compared to CLIP using the same GPU resource. 
Specifically, we can train DisCo-CLIP models with batch size 32,768 using 8 A100 40GB GPUs, or with batch size 196K using 64 GPUs.
We hope DisCo-CLIP will also help and motivate self-supervised learning~\cite{simclr_chen2020simple, mocov3_chen2021empirical} which also uses contrastive learning.

\section{Acknowledgement}
This work was partially supported by OPPO Research Fund.

%\input{cvpr2023-author_kit-v1_1-1/latex/introduction}
%\input{cvpr2023-author_kit-v1_1-1/latex/related_works}
%\input{cvpr2023-author_kit-v1_1-1/latex/algorithm}
%\input{cvpr2023-author_kit-v1_1-1/latex/experiments}

%%%%%%%%% REFERENCES
{\small
\bibliographystyle{ieee_fullname}
\bibliography{egbib}
}

\section{Appendix}
\subsection{Training Curves}

\begin{figure*}[t]
  \centering
   \includegraphics[width=0.48\linewidth]{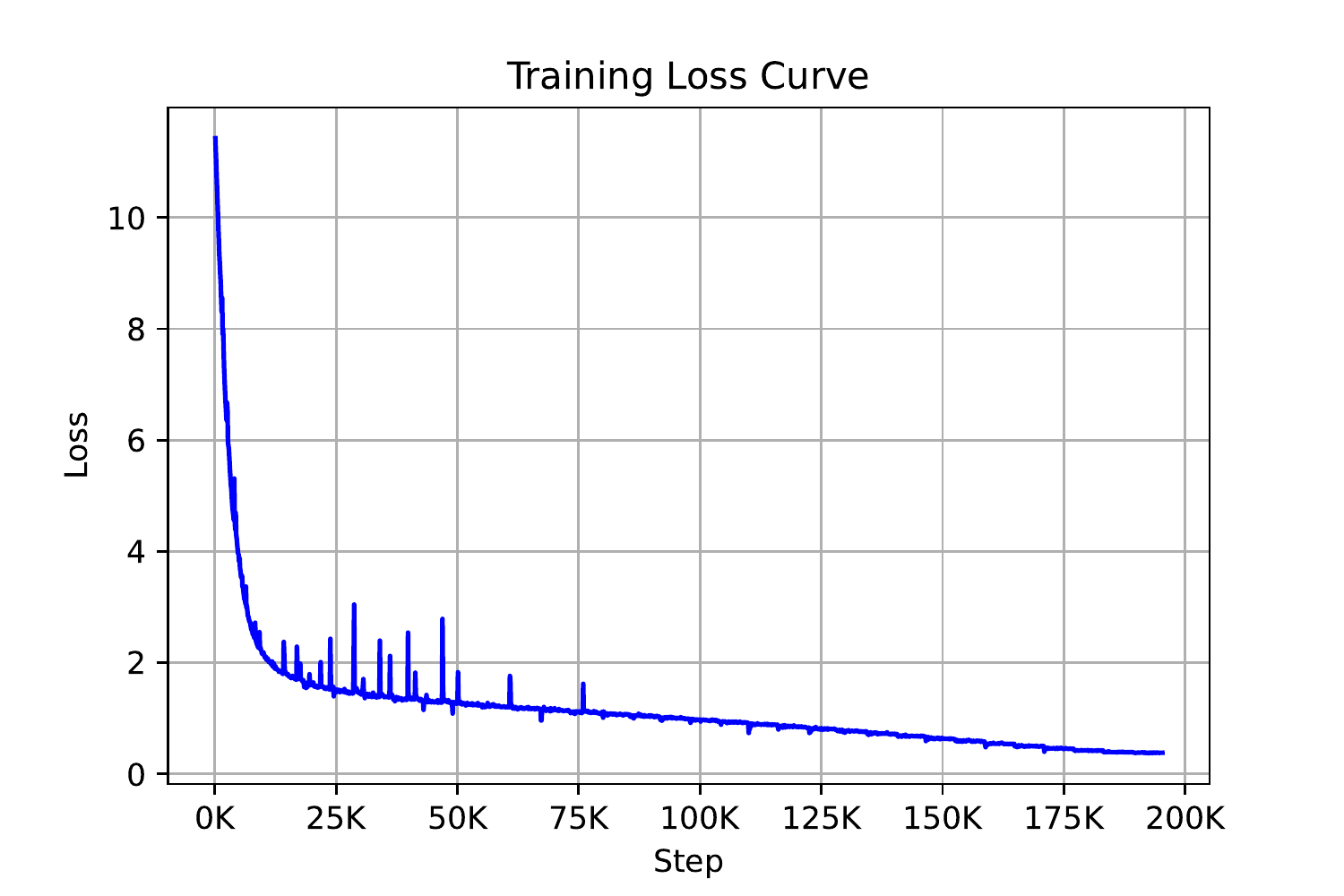}
   \includegraphics[width=0.48\linewidth]{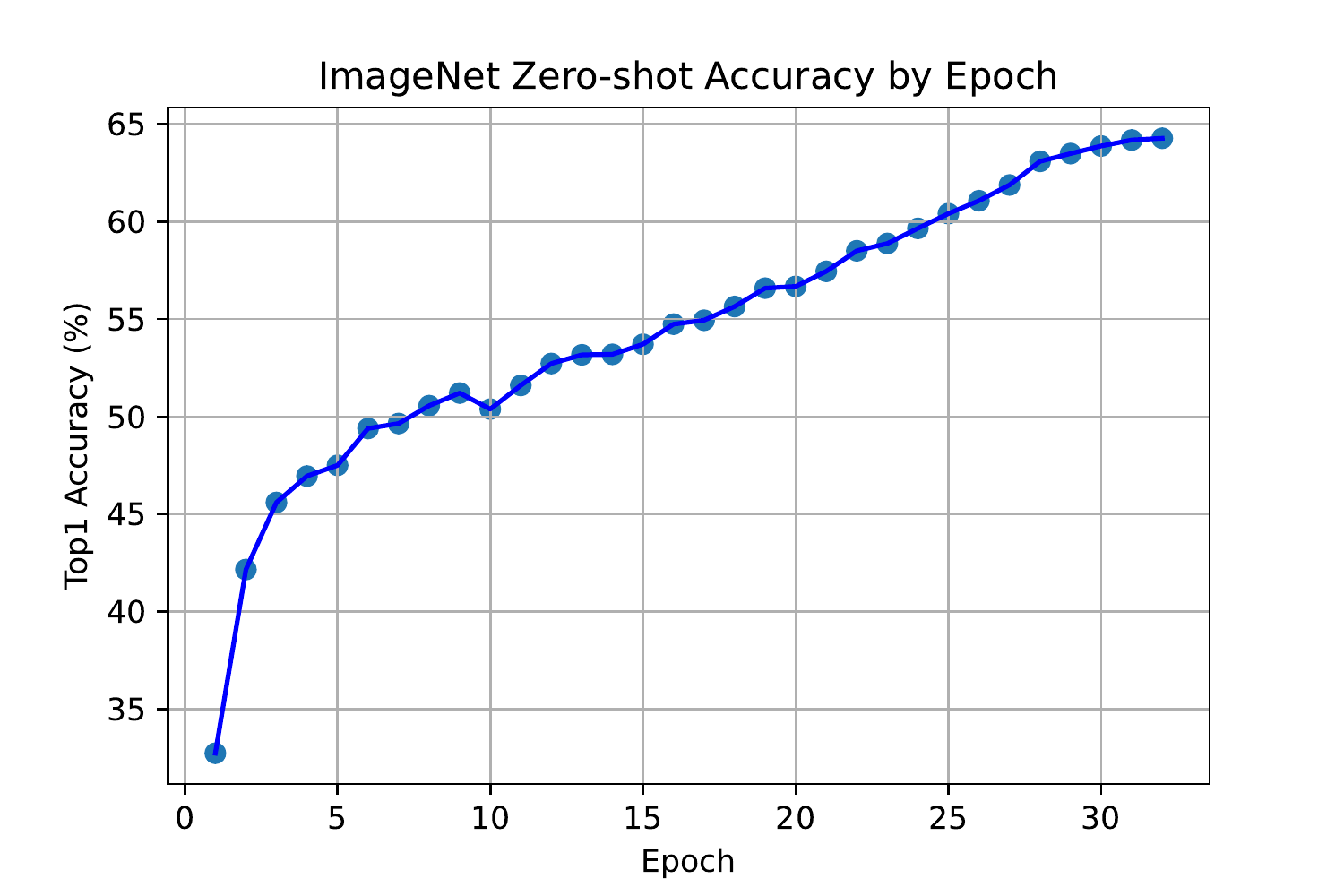}
   \caption{Curve of the training loss on the left and curve of the zero-shot top-1 ImageNet classification accuracy (\%) on the right. The base model is ViT-B/32, the used batch size  is 65,536. The model is trained 32 epochs on LAION-400M.}
   \label{fig:trainingcurves}
\end{figure*}

To facilitate the readers to observe the training process, we show the curves of the training loss and the zero-shot top-1 ImageNet classification in Fig.~\ref{fig:trainingcurves}. The base model is ViT-B/32, the used batch size  is 65,536. The model is trained 32 epochs on LAION-400M.

\subsection{Zero-shot Classification Details}

We follow CLIP~\cite{clip_radford2021learning} for the zero-shot  classification. We use the same 80 prompts as in CLIP. During evaluation, We first resize the image to the short side 224 according to the image ratio and then centrally crop a $224\times 224$ image.

\begin{table}[ht]
\small
\centering
\begin{tabular}{lcc}
\hline Dataset  & Classes & Number of Test Images \\ 
\hline 
INet~\cite{imagenet_deng2009imagenet}        & 1,000 & 50,000\\ 
INet-v2~\cite{imagenetv2_recht2019imagenet}     & 1,000 & 10,000 \\ 
INet-R~\cite{imagenetr_hendrycks2021many}      & 1,000 & 30,000 \\ 
INet-S~\cite{imagenets_wang2019learning} & 1,000 & 50,889 \\
\hline
\end{tabular}
\caption{Information of the evaluation data sets used in our paper.}
\label{tab:datasetintro}
\end{table}

Our used evaluation datasets include INet~\cite{imagenet_deng2009imagenet}, INet-v2~\cite{imagenetv2_recht2019imagenet}, INet-R~\cite{imagenetr_hendrycks2021many} and INet-S~\cite{imagenets_wang2019learning}. We show some basic information in Tab.~\ref{tab:datasetintro}. More information can be found in Timm\footnote{\url{https://github.com/rwightman/pytorch-image-models}}.

\subsection{More Experiments}
We further evaluate the performance of DisCo-CLIP under different batch sizes. The experiments are conducted on LAION-100M subset, the models are based on ViT-B/32, and the number of training epochs is 16.  The results are shown in Tab.~\ref{tab:moreexperiments}. We can see that from Tab.~\ref{tab:moreexperiments} larger batch size always brings in performance gain. 
Using DisCo-CLIP, it takes around 2 days to finish training of 16 epochs of LAION-100M with batch size 32,768 on a cluster with only 8 A100 GPUs. 

\begin{table*}
\small
    \centering
    \begin{tabular}{ccccccccc}
    \hline 
    Model & Datasets & Epochs & Steps & Batch Size & INet~\cite{imagenet_deng2009imagenet}  & INet-V2~\cite{imagenetv2_recht2019imagenet}  & INet-R~\cite{imagenetr_hendrycks2021many}  & INet-S~\cite{imagenets_wang2019learning}  \\ 
    DisCo-CLIP & LAION-100M$^{*}$ &16 & \({200} \mathrm{~K}\)  & 8,192 & \({48.76}\) & \({41.28}\) & {56.67} & \({36.65}\)  \\ 
    DisCo-CLIP & LAION-100M$^{*}$ &16 & \({100} \mathrm{~K}\)  & 16,384 & \({50.95}\) & \({43.07}\) & {59.24} & \({38.74}\)  \\ 
    DisCo-CLIP & LAION-100M$^{*}$ &16 & \({50} \mathrm{~K}\)  & 32,768 & \({51.64}\) & \({43.85}\) & {60.07} & \({39.25}\)  \\ 
    DisCo-CLIP & LAION-100M$^{*}$ &16 & \({25} \mathrm{~K}\)  & 65,536 & \(\textbf{51.91}\) & \(\textbf{44.19}\) & \textbf{60.52} & \(\textbf{39.76}\)  \\ 
    \hline
    \end{tabular}
    \caption{Performance evaluation of DisCo-CLIP under different batch size. We report top-1 zero-shot classification accuracy (\%)  on several data sets. All models are based on ViT-B/32. Our LAION-100M$^{*}$ is a 100M subset of LAION-2B.}
    \label{tab:moreexperiments}
\end{table*}

\end{document}